\def\year{2022}\relax
\documentclass[letterpaper]{article} 
\usepackage{aaai22}  
\usepackage{times}  
\usepackage{helvet}  
\usepackage{courier}  
\usepackage[hyphens]{url}  
\usepackage{graphicx} 
\usepackage{natbib}  
\usepackage{caption} 
\DeclareCaptionStyle{ruled}{labelfont=normalfont,labelsep=colon,strut=off} 
\usepackage{pgfplots}

\usepackage{algorithm}
\usepackage{algorithmic}
\usepackage{booktabs}
\usepackage{amssymb}
\usepackage{pifont}
\usepackage{multicol}
\usepackage{multirow}
\usepackage{tikz}
\usepackage{amsmath}

\usepackage{newfloat}
\usepackage{listings}
\floatstyle{ruled}
\newfloat{listing}{tb}{lst}{}
\floatname{listing}{Listing}

\DeclareMathAlphabet\mathbfcal{OMS}{cmsy}{b}{n}
\makeatletter
\def\@fnsymbol#1{\ensuremath{\ifcase#1\or \dagger\or *\or \ddagger\or
   \mathsection\or \mathparagraph\or \|\or **\or \dagger\dagger
   \or \ddagger\ddagger \else\@ctrerr\fi}}
   
\title{Understanding and Improving the Exemplar-based Generation \\ for Open-domain Conversation}
\author{
Seungju Han\thanks{Equal contribution.}\quad Beomsu Kim\footnotemark[1]\quad Seokjun Seo\footnotemark[1]\quad Enkhbayar Erdenee\footnotemark[1]\quad Buru Chang\thanks{Corresponding author.}\\
Hyperconnect\\
{\tt \normalsize \{seungju.han,beomsu.kim,seokjun.seo,enkhbayar.erdenee,buru.chang\}@hpcnt.com}
}

\usepackage{bibentry}

\DeclareMathAlphabet\mathbfcal{OMS}{cmsy}{b}{n}

\begin{document}

\maketitle

\begin{abstract}\label{sec:0_abstract}
Exemplar-based generative models for open-domain conversation produce responses based on the exemplars provided by the retriever, taking advantage of generative models and retrieval models.
However, they often ignore the retrieved exemplars while generating responses or produce responses over-fitted to the retrieved exemplars.
In this paper, we argue that these drawbacks are derived from the one-to-many problem of the open-domain conversation.
When the retrieved exemplar is relevant to the given context yet significantly different from the gold response, the exemplar-based generative models are trained to ignore the exemplar since the exemplar is not helpful for generating the gold response.
On the other hand, when the retrieved exemplar is lexically similar to the gold response, the generative models are trained to rely on the exemplar highly.
Therefore, we propose a training method selecting exemplars that are semantically relevant to the gold response but lexically distanced from the gold response to mitigate the above disadvantages.
In the training phase, our proposed training method first uses the gold response instead of dialogue context as a query to select exemplars that are semantically relevant to the gold response.
And then, it eliminates the exemplars that lexically resemble the gold responses to alleviate the dependency of the generative models on that exemplars.
The remaining exemplars could be irrelevant to the given context since they are searched depending on the gold response.
Thus, our proposed training method further utilizes the relevance scores between the given context and the exemplars to penalize the irrelevant exemplars.
Extensive experiments demonstrate that our proposed training method alleviates the drawbacks of the existing exemplar-based generative models and significantly improves the performance in terms of appropriateness and informativeness.
\end{abstract}
\section{Introduction}\label{sec:1_introduction}
Exemplar-based generative models~\cite{wu2019response,weston2018retrieve,cai2019retrieval,gupta2021controlling} for open-domain conversation combine a retrieval model~\cite{humeau2019poly,mazare2018training,kim2021distilling} and a generative model~\cite{adiwardana2020towards,roller2021recipes,zhang2020dialogpt,NEURIPS2020_1457c0d6} into a single framework to generate responses in two steps:
(1) the retriever searches an exemplar using the given context as a query, and (2) the generator produces a response based on the given context and the retrieved exemplar.
Exemplar-based generative models produce more specific responses than those produced by vanilla generative models while being more fluent than responses searched by retrieval models.
\begin{figure}[t]
\centering
\includegraphics[width=0.98\columnwidth]{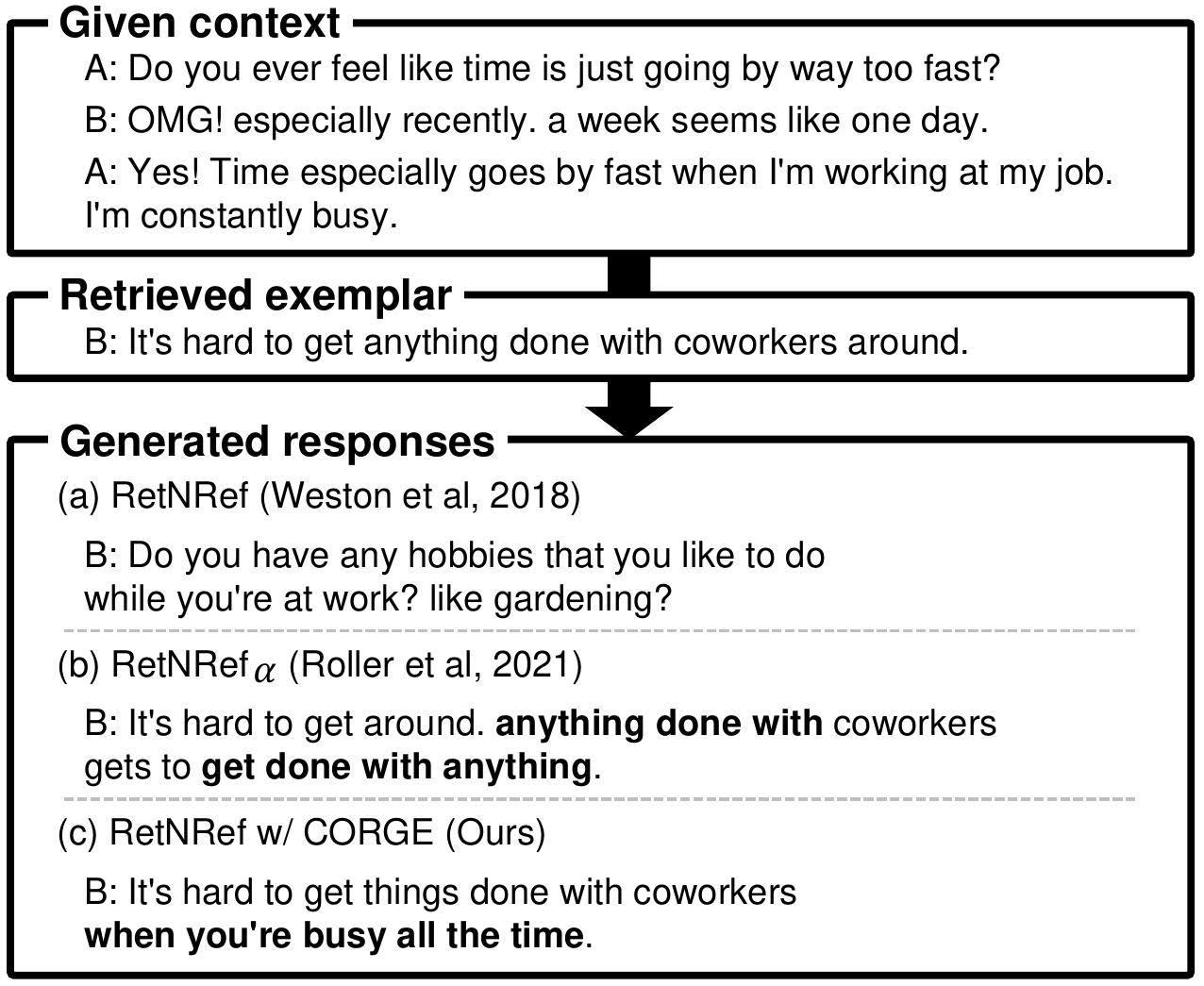}
\caption{
Responses generated by the three exemplar-based generative models.
\textit{RetNRef} ignores the exemplar during response generation, \textit{RetNRef$_\alpha$} generates the response highly over-fitted to the exemplar, and \textit{RetNRef} trained with our proposed training method (CORGE) well utilizes the exemplar to produce a more fluent response than that of the others.}
\vspace*{-1.0em}
\label{fig:1_exmplar_based_generation_examples}
\end{figure}

Despite their success, exemplar-based generative models have two major shortcomings.
Primitive exemplar-based generative models~\citep{weston2018retrieve,cai2019skeleton} tend to \textbf{\textit{entirely ignore the exemplars}} and produce responses similar to those of vanilla generative models.
This is due to the \textit{one-to-many problem}~\citep{li2016diversity} where there are many possible responses for each dialogue context.
During the training phase, the retrieved exemplar is not helpful for generating the gold response when the exemplar retrieved for the given context is significantly different from the gold response.
This leads exemplar-based generative models to ignore the exemplar while generating responses, as shown in Figure~\ref{fig:1_exmplar_based_generation_examples}(a).
To address this issue, recent exemplar-based generative models utilize the gold response~\citep{roller2021recipes} or the slightly perturbed gold response~\citep{cai2019retrieval} as an exemplar in the training phase.
However, these training methods cause the generator to \textbf{\textit{rely heavily on the retrieved exemplar}}, i.e. the generator resorts to copying the provided tokens, as shown in Figure~\ref{fig:1_exmplar_based_generation_examples}(b).
These two disadvantages of existing exemplar-based generative models can adversely affect the quality of the generated response.

Therefore, we propose \textbf{\textit{CORGE}} (COnnecting Retriever and GEnerator), a simple training method of exemplar-based generative models considering the one-to-many problem of the open-domain conversation.
As inspired by \citeauthor{wu2019response}~\shortcite{wu2019response}, CORGE first utilizes the gold response instead of dialogue context as the query for the retriever to select exemplars that are similar to the gold response.
The retrieved exemplars ensure that exemplar-based generative models utilize their semantics while generating the gold response at the training phase.
Since the exemplars are retrieved by the gold response, some of them are lexically identical or too similar to the gold response.
These exemplars lead exemplar-based generative models to be trained to depend on the exemplar heavily.
Thus, CORGE then eliminates the exemplars based on the distance between the exemplars and the gold response to alleviate the dependency of the generative models on the exemplars.
Here, we employ Jaccard similarity to measure the distance \cite{guu2018generating,cai2019skeleton,wu2019response}.
However, as the selected exemplars solely depend on the gold response, some of them may be irrelevant to the given context, which results in exemplar-based generative models still ignoring the retrieved exemplar.
To solve this, CORGE utilizes the relevance scores between the context and the exemplar to weight the relevant exemplars and penalizes irrelevant exemplars to the given context.
Extensive experiments show that CORGE is generally applicable to the existing exemplar-based generative models and improves the quality of generated responses regarding appropriateness and informativeness.

Our main contributions are summarized as follows:
\begin{itemize}
  \item We analyze the shortcomings of existing exemplar-based generative models derived from the nature of the open-domain conversation, the one-to-many problem.
  \item We propose a training method (CORGE) to improve the quality of generated responses by selecting useful exemplars and weighting the exemplars by relevance scores assessed by the retriever.
  \item Through the human evaluation, we demonstrate that CORGE significantly improves the performance of exemplar-based generative models in terms of appropriateness and informativeness.
\end{itemize}
\begin{figure*}[t]
\centering
\includegraphics[width=0.90\textwidth]{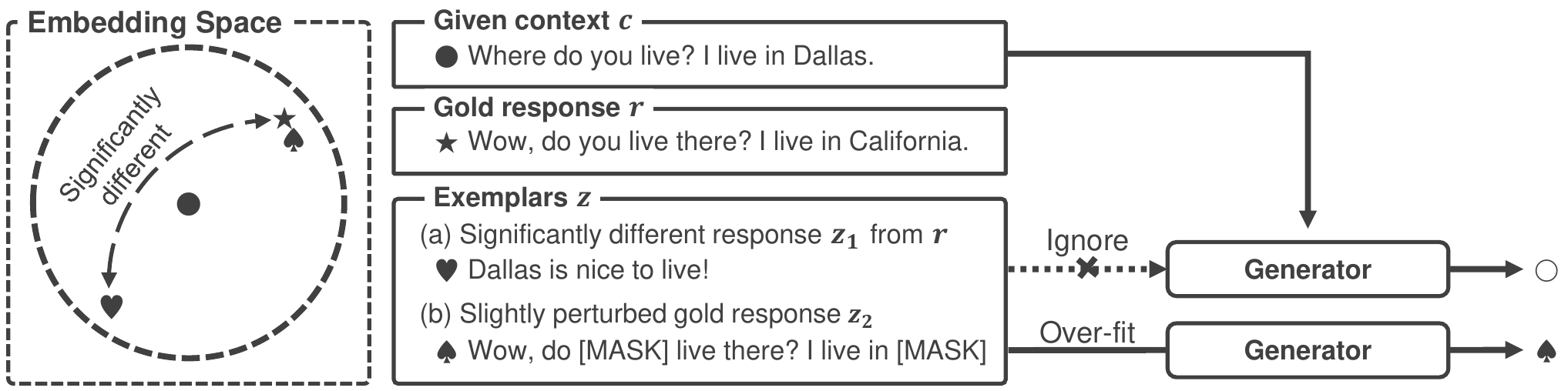}
\caption{Illustration of the drawbacks of existing exemplar-based generative models. The black dotted line indicates the boundary of the relevant exemplars to the given context.}
\vspace*{-1em}
\label{fig:2_drawbacks}
\end{figure*}

\section{Related Work}\label{sec:2_related_work}
\subsection{Exemplar-based Generation}\label{subsec:2_1_related_work_open_domain_conversation}
While generative models have shown remarkable performance on the open-domain conversation, it is well-known that generative models tend to yield uninformative and bland responses~\cite{li2016diversity,liu2016not,serban2017multiresolution,li2020don,holtzman2019curious,welleck2019neural}. 
Exemplar-based generative models are introduced to overcome the aforementioned problem generative models suffer.
\citet{wu2019response} introduce an exemplar-based generative model for open-domain conversation, which retrieves a context-exemplar pair conditioned by the input context and encodes the lexical difference between the input context and the retrieved context to the edit vector.
The response is produced by feeding the exemplar and the edit vector to the generator.
\citet{weston2018retrieve,roller2021recipes} also retrieve the exemplar using the given context as a query and concatenate the exemplar with the context, then feed the concatenated exemplar into the generator to produce the final response for the open-domain conversation.
\citet{cai2019skeleton,cai2019retrieval} propose a method that removes the irrelevant information from the exemplar, then uses the masked exemplar to inform the generator to produce the response.
\citet{gupta2021controlling} condition the generator with the retrieved exemplars and the extracted semantic frames of the exemplars, which improves the coherence of generated responses.
We do not consider this model as a baseline because their model requires an additional semantic frame extractor, and it can be mutually complemented with our proposed training method.

\subsection{Knowledge-grounded Generation}\label{subsec:2_2_related_work_rag}
Knowledge-grounded generation models that utilize retrieved results (e.g., relevant documents from Wikipedia) to generate informative responses have been proposed to perform knowledge-intensive NLP tasks (e.g., open-domain question answering).
The knowledge-grounded generation has a similar form with the exemplar-based generation.
However, the main difference is that knowledge-grounded generative models extract the knowledge from external resources to generate the informative response.
\citet{guu2020realm} show the effectiveness of pre-training a knowledge retriever with the large-scale language model for open-domain question answering, and \citet{NEURIPS2020_6b493230} demonstrate that knowledge-grounded generative models produce more informative and diverse sentences than vanilla generative models on a wide range of knowledge-intensive NLP tasks.
\citet{fan2021augmenting} similarly propose a knowledge-grounded generative model for response generation, but they do not focus on the open-domain conversation.
In \textit{Method Section}, we demonstrate the difference between our approach and knowledge-grounded generative models, and we show that existing knowledge-grounded generative models are not directly applicable to the open-domain conversation in \textit{Experiments Section}.
\section{Preliminaries}\label{sec:3_preliminaries}
In this section, we first introduce the exemplar-based generation with notations, and then describe the drawbacks of prior studies.

\subsection{Exemplar-based Generation}\label{subsec:3_1_method_task_formulation}
Let $D=\{(c_i, r_i) \mid 1 \leq i \leq n\}$ denote the dialogue dataset, which consists of $n$ pairs of context $c$ and response $r$.
Exemplar-based generative models are composed of two components: a retriever $\mathbfcal{R}$ and a generator $\mathbfcal{G}$.
For a given context $c_i$, the retriever finds the top-scoring exemplar based on the relevance score $S_{\mathbfcal{R}}(z, c_i)$ of the exemplar $z \in R$ , where $R$ is a pre-defined response set.
The generator computes the probability of the response for the context $c_i$ while utilizing the exemplar $z$ as $P_{\mathbfcal{G}}(r | c_i, z)$.

\subsection{Drawbacks of Existing Exemplar-based Generative models}\label{subsec:3_2_drawbacks_of_existing_exemplar_based_generatinve_models}
As mentioned in~\citet{roller2021recipes}, the primitive exemplar-based generative model~\citep{weston2018retrieve} tends to ignore the retrieved exemplar during response generation due to the one-to-many problem in open-domain conversation~\citep{li2016diversity}.
Since its retriever searches an exemplar based on a given context, the retrieved exemplar is often significantly different from a gold response of the generator, although both of the retrieved exemplar and gold response are relevant to the given context, which is shown in Figure~\ref{fig:2_drawbacks}(a).
As the retrieved exemplar is not helpful for generating the gold response, the generator is trained to ignore the retrieved exemplar and to produce a response using only the given context.

To induce the generator to utilize retrieved exemplars more actively, \citet{roller2021recipes} make use of the gold response, and \citet{cai2019retrieval} use perturbed gold response as an exemplar rather than using retrieved exemplars during the model training.
However, since the exemplar $z_i$ and the gold response $r_i$ are too similar (as shown in Figure~\ref{fig:2_drawbacks}(b)), the exemplar-based generative model learns to rely overly on the exemplar.
Eventually, the generator produces a highly over-fitted response to the exemplar by directly copying the tokens of the exemplar.
\section{Method}\label{sec:4_method}
\begin{figure}[t]
\centering
\includegraphics[width=\columnwidth]{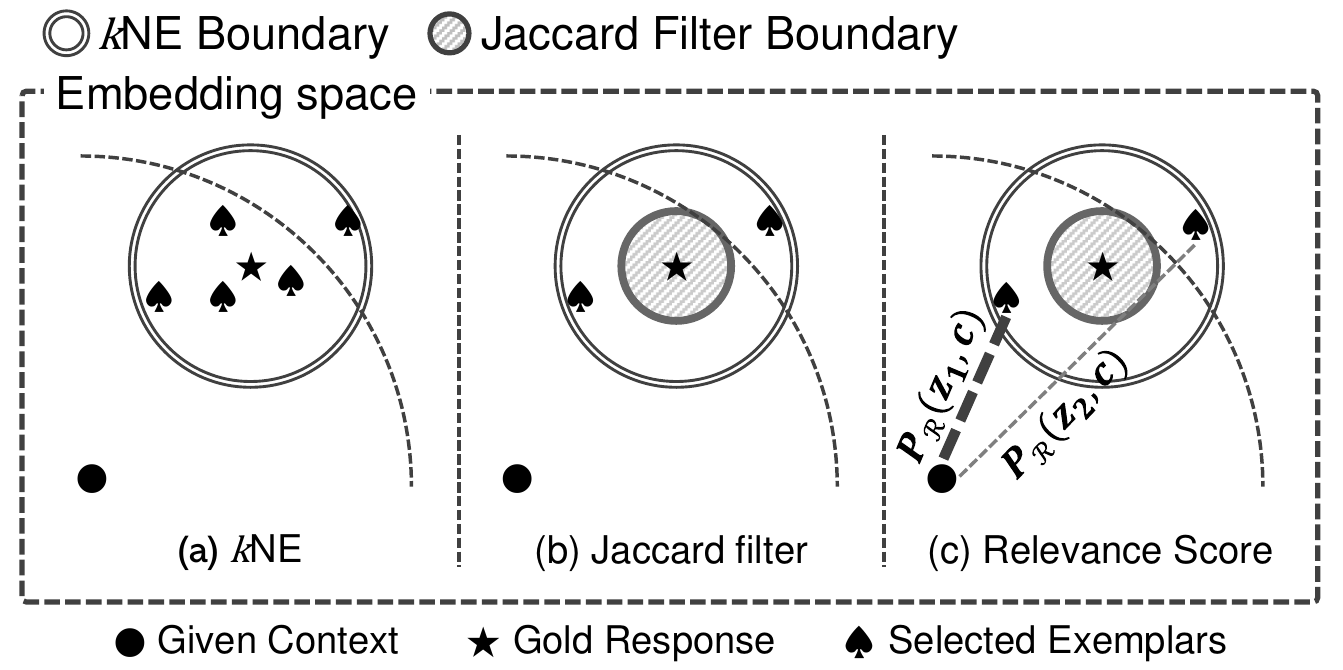}
\caption{The procedure of our proposed training method, CORGE. (a): Selecting $k$NE of the gold response $r$ based on $S_{\mathbfcal{R'}}(z, r)$. (b): Filtering out the exemplars which are too close to the gold response $r$. (c): Weighting the exemplars $z$ depending on their normalized relevance scores $P_{\mathbfcal{R}}(z, c)$.}
\vspace*{-0.8em}
\label{fig:3_bridge}
\end{figure}

We hypothesize that selecting semantically relevant but lexically distanced exemplars from the gold response could solve the drawbacks above.
To validate this hypothesis, we introduce a training method of exemplar-based generative models, called CORGE.
Our proposed training method is illustrated in Figure~\ref{fig:3_bridge}, and the illustrative examples about the exemplars selected by CORGE are described in Table~\ref{tab:1_exemplar_sample}.

\subsection{Selecting Exemplars Semantically Relevant but Lexically Distanced to the Gold Response}\label{subsec:4_1_finding_proper_exemplars}
We describe how CORGE selects semantically relevant but lexically distanced exemplars to the gold response.
Conventionally, the retriever selects the exemplars $z$ based on the relevance score $S_{\mathbfcal{R}}(z, c_i)$ for the given context $c_i$.
However, this searching process could return a significantly different exemplar $z$ from the gold response $r_i$, and it induces the generator $\mathbfcal{G}$ to ignore the retrieved exemplar during response generation.
Therefore, we select exemplars based on the gold response $r_i$ to ensure that the generator $\mathbfcal{G}$ utilizes the exemplars inspired by \citeauthor{wu2019response}.
We select top-$k$ scoring exemplars based on the score $S_{\mathbfcal{R'}}(z, r_i)$, which we call \textit{$k$-Nearest Exemplars ($k$NE)}.\footnote{Note that $S_\mathbfcal{R}(z,c)$ and $S_\mathbfcal{R'}(z,r_i)$ use the same retriever, but they are computed differently. Please refer to how we calculate the score $S_{\mathbfcal{R'}}(z, r_i)$ and $S_{\mathbfcal{R}}(z, c)$ in the Supplementary Materials.}
These $k$NE are more semantically related to the gold response $r_i$ than the exemplar obtained by using $S_{\mathbfcal{R}}(z, c_i)$.

However, some of the selected $k$NE are lexically identical or too close to the gold response $r$ unintentionally since the retriever searches the exemplars based on the gold response.
We observe that using these exemplars also causes the over-fitting problem of generated responses; therefore, the generator excessively copies tokens from the exemplars.
From this, we are motivated to filter out the exemplars which are lexically too close to the gold response and preserve the exemplars properly distanced to the gold response to mitigate the over-fitting problem.
Here, we employ \textit{Jaccard similarity} to measure the lexical similarity~\citep{guu2018generating,cai2019skeleton,wu2019response} between the exemplar and the gold response.
Exemplars are filtered out when their Jaccard distance with the gold response $r$ is larger than $0.6$, and we replace them with the randomly chosen responses from the pre-defined response set $R$.
The threshold of filtering is empirically chosen as 0.6.
The set of the final exemplars $z$ obtained through these steps is referred to as $Z_i = \{z_{i,1}, z_{i,2}, \cdots, z_{i,k}\}$.

\begin{table}[t]
\centering
\footnotesize
\setlength\tabcolsep{1.9pt}
\begin{tabular}{ccc}
\toprule
\multicolumn{3}{c}{\textbf{Input Context}}\\
\midrule
\multicolumn{3}{l}{\begin{tabular}[l]{l}What kind of animals you take care of?\end{tabular}} \\ 
\midrule
\multicolumn{3}{c}{\textbf{Gold Response}} \\
\midrule
\multicolumn{3}{l}{\begin{tabular}[l]{l}I work with a variety of animals. I sometimes work \\ with lions and monkeys.\end{tabular}} \\ \midrule
\midrule
\multicolumn{1}{c}{\textbf{Context Retrieval}} &
\multicolumn{1}{c}{\textbf{Sim}} & \multicolumn{1}{c}{$P_{\mathbfcal{R}}(z,c)$} \\
\midrule
\multicolumn{1}{l}{\begin{tabular}[l]{l}I raise two dogs.\end{tabular}}  & \multicolumn{1}{c}{0.1} & \multicolumn{1}{c}{0.9}\\
\midrule
\multicolumn{1}{c}{\textbf{\textit{k}NE}} & \multicolumn{1}{c}{\textbf{Sim}} & \multicolumn{1}{c}{$P_{\mathbfcal{R}}(z, c)$} \\
\midrule
\multicolumn{1}{l}{\begin{tabular}[l]{l}I work with a variety of animals.\end{tabular}}  & \multicolumn{1}{c}{\textbf{0.9}} &  \multicolumn{1}{c}{0.2}\\
\midrule
\multicolumn{1}{l}{\begin{tabular}[l]{l}He works with various people.\end{tabular}}  & \multicolumn{1}{c}{0.3} & \multicolumn{1}{c}{\textbf{0.0}}\\
\midrule
\multicolumn{1}{l}{\begin{tabular}[l]{l}I work with lots of different animals.\end{tabular}}  & \multicolumn{1}{c}{0.5} & \multicolumn{1}{c}{0.3}\\
\midrule
\multicolumn{1}{l}{\begin{tabular}[l]{l}I like work with animals a lot, \\sometimes for work or fun.\end{tabular}}  & \multicolumn{1}{c}{0.5} & \multicolumn{1}{c}{0.2}\\
\midrule
\multicolumn{1}{l}{\begin{tabular}[l]{l}I do some work with animals they’re \\amazing creatures.\end{tabular}}  & \multicolumn{1}{c}{0.3} & \multicolumn{1}{c}{0.3}\\
\bottomrule
\end{tabular}
\caption{Samples of the exemplars selected by CORGE.
\textbf{Context Retrieval} indicates the exemplar retrieved by using the context as a query, and \textbf{\textit{k}NE} shows the exemplars selected by using the gold response as a query.
\textbf{Sim} measures the lexical similarity between the gold response and the exemplar and $P_{\mathbfcal{R}}(z,c)$ indicates the normalized relevance score calculated by retriever.
}

\vspace*{-0.5em}
\label{tab:1_exemplar_sample}
\end{table}

\subsection{Weighting the Selected Exemplars based on the Relevance Score}
As we select the exemplar totally based on the gold response, some of $k$NE could be relevant to the gold response $r_i$ but irrelevant to the given context $c_i$.
Therefore, we condition the generator with the relevance score of $k$NE to reward the relevant exemplars and penalize irrelevant exemplars.
Using the retriever $\mathbfcal{R}$, we calculate the relevance score $S_{\mathbfcal{R}}(z_{i,j}, c_i)$ per each selected exemplar $z_{i,j}$, then apply the softmax function to the relevance score to obtain the normalized relevance score $P_\mathbfcal{R}(z_{i,j}, c_i)$.
Then we replace the traditional likelihood with the weighted likelihood using the normalized score.
Our final training objective is to minimize the loss function $L = \sum_{i=1}^{n}L(r_i, c_i)$ where:
\begin{equation}
L(r_i, c_i) = -\log{\sum_{z \in Z_i}P_{\mathbfcal{R}}(z, c_i)P_\mathbfcal{G}(r_i|c_i,z)}
\end{equation}

The gradient of the generator $\mathbfcal{G}$ is calculated as follows:
\begin{equation}
\nabla_\mathbfcal{G} L(r_i, c_i) = -\alpha \cdot \sum_{z \in Z_i}P_{\mathbfcal{R}}(z, c_i) \nabla_\mathbfcal{G}(P_{\mathbfcal{G}}(r_i|c_i,z)),
\end{equation}
where $\alpha^{-1} = \sum_{z\in Z_i}P_{\mathbfcal{R}}(z, c_i)P_{\mathbfcal{G}}(r_i|c_i,z)$. 
This equation demonstrates that the gradient of the generator $\mathbfcal{G}$ is scaled by the normalized relevance score $P_\mathbfcal{R}(z, c_i)$, which indicates that the generator is less updated when the retrieved exemplar $z$ is not relevant to the given context $c_i$.
This procedure helps the model ignore the irrelevant exemplars.
Thus, the generator learns to fetch tokens from the exemplar more easily, which is relevant to the gold response.

\noindent
\textbf{Difference between CORGE and Knowledge-grounded generative models}
The way of leveraging the relevance scores is already employed by knowledge-grounded generative models~\cite{NEURIPS2020_6b493230,sachan2021end} in open-domain question answering.
However, there is a significant difference between our CORGE and knowledge-grounded generative models.
CORGE uses the relevance score $P_{\mathbfcal{R}}(z, c_i)$ to penalize the irrelevant exemplars $z$ to the given context $c_i$ since the exemplars are retrieved by $S_{\mathbfcal{R'}}(z, r_i)$.
Knowledge-grounded generative models use it as the latent variable to jointly train the retriever $\mathbfcal{R}$ and generator $\mathbfcal{G}$.
Especially, knowledge-grounded generative models also tend to ignore the retrieved exemplars due to the one-to-many nature in open-domain conversation when the retriever and generator are jointly trained.
On the other hand, we do not perform the joint learning of the retriever and the generator, but freeze the retriever while training the generator.
\section{Experiments}\label{sec:5_experiments}
\subsection{Dataset}\label{subsec:3_1_training_data}
We utilize the following four datasets used in \citet{roller2021recipes}, which are Blended Skill Talk (BST)~\citep{smith2020can}, ConvAI2~\citep{zhang2018personalizing}, Empathetic Dialogues (ED)~\citep{rashkin2019towards}, and Wizard of Wikipedia (WoW)~\citep{dinan2018wizard}.
To simplify the notation, we denote the concatenated version of these four datasets as \textbf{\textit{BST+}}.
We split BST$+$ into train, validation, and test sets following~\citet{smith2020can}.

\subsection{Baselines}\label{subsec:3_2_baselines}
\paragraph{Retrieval and Generative Models}
\textit{Bi-encoder 256M}~\citep{mazare2018training} and \textit{Blender 90M}~\cite{roller2021recipes} are considered as a baseline retrieval model and a baseline generative model.
Further, they are also employed as a retriever and a generator of the following exemplar-based generative baselines, respectively.

\paragraph{Exemplar-based Generative Models}
Since our proposed training method is for training exemplar-based generation models,  we first consider recent exemplar-based generation models,
\textit{RetNRef}~\citep{weston2018retrieve}, \textit{RetNRef$_\alpha$}~\citep{roller2021recipes}, and \textit{MatToGen}~\citep{cai2019retrieval}, as baselines.
\textit{RetNRef} concatenates the retrieved exemplar with the given context as the input of the generator to produce the response.
\textit{RetNRef$_\alpha$} is the dialogue retrieval version of \textit{RetNRef}, which adopts  $\alpha$-blending to escape from simply ignoring the retrieved exemplars ($\alpha=0.5$).
\textit{MatToGen} extracts the meaningful tokens from the exemplar to provide them to the generator.

To verify the effectiveness of our training method, we apply CORGE to \textit{RetNRef} and \textit{MatToGen} instead of their training method.
They are denoted as \textit{RetNRef}+\textit{CORGE} and \textit{MatToGen}+\textit{CORGE}, respectively.

\paragraph{Knowledge-grounded Generative Models}
Although \textit{RAG}~\citep{NEURIPS2020_6b493230} and \textit{KIF}~\cite{fan2021augmenting} are proposed to perform knowledge-grounded generation tasks, we employ \textit{RAG} and \textit{KIF} as baselines since they have a similar form with exemplar-based generative models. 
Our experiments demonstrate that these knowledge-grounded generative models cannot be directly applied to the open-domain conversation.

\begin{table*}[ht]
\footnotesize
\centering
\setlength\tabcolsep{3.0pt}
\begin{tabular}{lcccccccc}
\toprule
\multicolumn{1}{c}{\multirow{2}{*}{\textbf{Model Names (A vs. B)}}} & \multicolumn{4}{c}{\textbf{Appropriateness (\%)}} & \multicolumn{4}{c}{\textbf{Informativeness (\%)}} \\
\cmidrule(lr){2-5} \cmidrule(lr){6-9}
\multicolumn{1}{c}{} &
\multicolumn{1}{c}{\textbf{Win Rate}} &
\multicolumn{1}{c}{\textbf{A win}} & 
\multicolumn{1}{c}{\textbf{Tie}} & 
\multicolumn{1}{c}{\textbf{B win}} &
\multicolumn{1}{c}{\textbf{Win Rate}} &
\multicolumn{1}{c}{\textbf{A win}} & 
\multicolumn{1}{c}{\textbf{Tie}} & 
\multicolumn{1}{c}{\textbf{B win}} \\
\midrule
RetNRef$_\alpha$ vs. Bi-encoder 256M
& 44.9$\:\;$ & 32.0$\:\;$ & 28.7$\:\;$ & \textbf{39.3}$\:\;$ & 47.5$\:\;$ & 31.3$\:\;$ & 34.0$\:\;$ & \textbf{34.7}$\:\;$ \\
RetNRef$_\alpha$ vs. Blender 90M
& 50.2$\:\;$ & \textbf{37.3}$\:\;$ & 25.7$\:\;$ & 37.0$\:\;$  & 53.3$\:\;$ & \textbf{40.3}$\:\;$ & 24.3$\:\;$ & 35.4$\:\;$ \\
\midrule
RetNRef + CORGE vs. Bi-encoder 256M
& 52.6$\:\;$ & \textbf{34.0}$\:\;$ & 35.3$\:\;$ & 30.7$\:\;$ & 51.9$\:\;$ & \textbf{35.7}$\:\;$ & 31.3$\:\;$ & 33.0$\:\;$ \\
RetNRef + CORGE vs. Blender 90M
& 57.7$^*$ & \textbf{33.7$^*$} & 41.7$^*$ & 24.6$^*$ & 54.6$\:\;$ & \textbf{30.0}$\:\;$ & 45.0$\:\;$ & 25.0$\:\;$ \\
RetNRef + CORGE vs. RetNRef$_\alpha$
& 53.2$\:\;$ & \textbf{30.3}$\:\;$ & 43.0$\:\;$ & 26.7$\:\;$ & 51.6$\:\;$ & \textbf{27.7}$\:\;$ & 46.3$\:\;$ & 26.0$\:\;$ \\
RetNRef + CORGE vs. RetNRef 
& 54.4$\:\;$ & \textbf{41.0}$\:\;$ & 24.7$\:\;$ & 34.3$\:\;$ & 53.4$\:\;$ & \textbf{37.0}$\:\;$ & 30.7$\:\;$ & 32.3$\:\;$\\
RetNRef + CORGE vs. KIF & 57.5$^*$ & \textbf{37.0}$^*$ & 35.7$^*$ & 27.3$^*$ & 50.0$\:\;$ & 30.0$\:\;$ & 40.0$\:\;$ & 30.0$\:\;$ \\
RetNRef + CORGE vs. RAG & 53.5$\:\;$ & \textbf{37.7}$\:\;$ & 29.7$\:\;$ & 32.6$\:\;$ & 52.1$\:\;$ & \textbf{29.7}$\:\;$ & 43.0$\:\;$ & 27.3$\:\;$ \\
\midrule\midrule
MatToGen vs. Bi-encoder 256M
& 47.1$\:\;$ & 33.3$\:\;$  & 29.3$\:\;$ & \textbf{37.4}$\:\;$ & 50.9$\:\;$ & \textbf{36.7}$\:\;$ & 28.0$\:\;$ & 35.3$\:\;$ \\
MatToGen vs. Blender 90M 
& 48.1$\:\;$ & 34.0$\:\;$  & 29.3$\:\;$ & \textbf{36.7}$\:\;$ & 46.3$\:\;$ & 31.6$\:\;$  & 31.7$\:\;$  & \textbf{36.7}$\:\;$ \\
\midrule
MatToGen + CORGE vs. Bi-encoder 256M
& 54.2$\:\;$ & \textbf{43.0}$\:\;$ & 20.7$\:\;$ & 36.3$\:\;$ & 54.4$\:\;$ & \textbf{41.3}$\:\;$ & 24.0$\:\;$ & 34.7$\:\;$  \\
MatToGen + CORGE vs. Blender 90M
& 58.0$^*$  & \textbf{35.0}$^*$ & 39.7$^*$ & 25.3$^*$ & 58.1$^*$  & \textbf{36.0}$^*$ & 38.0$^*$ & 26.0$^*$  \\
MatToGen + CORGE vs. MatToGen
& 52.6$\:\;$ & \textbf{33.3}$\:\;$ & 36.7$\:\;$ & 30.0$\:\;$  & 53.3$\:\;$ & \textbf{32.7}$\:\;$ & 38.7$\:\;$ & 28.6$\:\;$ \\
MatToGen + CORGE vs. KIF & 57.1$^*$ & \textbf{44.0}$^*$ & 23.0$^*$ & 33.0$^*$ & 52.5$\:\;$ & \textbf{39.0}$\:\;$ & 25.7$\:\;$ & 35.3$\:\;$ \\
MatToGen + CORGE vs. RAG  & 51.6$\:\;$ & \textbf{38.3}$\:\;$ & 25.7$\:\;$ & 36.0$\:\;$ & 55.6$\:\;$ & \textbf{41.3}$\:\;$ & 25.7$\:\;$ & 33.0$\:\;$\\
\bottomrule
\end{tabular}
\caption{Pair-wise human evaluation results show that our proposed training method improves the performance against the existing exemplar-based generation approaches in terms of appropriateness and informativeness.
The win rate is calculated by excluding the tie.
$^*$ indicates statistical significance (two-tailed binomial test, p $<$ 0.05).}
\vspace*{-0.5em}
\label{tab:2_pairwise}
\end{table*}
\subsection{Evaluation Metrics}\label{subsec:3_3_evaluation_metrics}
To assess the performance of open-domain conversation models, we conduct a pair-wise comparison through the human evaluation following~\citet{weston2018retrieve}.
We use two criteria: \textbf{Appropriateness} and \textbf{Informativeness}. 
Appropriateness measures how the generated response is fluent, logical, and appropriate to the given context.
Informativeness measures how the generated response has meaningful information relevant to the given context.
We use Amazon Mechanical Turk to collect the annotations, and more details are described in the Supplementary Material.
Note that we consider pair-wise comparison since the pair-wise comparison results are more robust than Likert scores~\cite{kulikov2019importance,liang2020beyond}.

We also employ the automatic evaluation metrics, \textbf{Perplexity} (PPL), \textbf{Dist-$n$}, and \textbf{BLEU}~\cite{papineni2002bleu}, to analyze the generated responses of each model. 
PPL measures how well the model predicts a response based on the given input context, and lower PPL indicates that the model predicts the response better.
To analyze how much the exemplar-based generative model leverages the retrieved exemplar, we introduce two variants of PPL by utilizing conditional probability when exemplars are given: 
(1) PPL$_{gold}$ uses the conditional probability $P_\mathbfcal{G}(r|c, r)$, which assumes the situation when the gold response is given as an exemplar, and (2) PPL$_{retrieve}$ uses the conditional probability  $P_\mathbfcal{G}(r|c, z)$ where $z$ is the retrieved exemplar by using $S_\mathbfcal{R'}(z, r)$.
Lower PPL$_{gold}$ denotes that the exemplar-based generative model predicts the gold response well when the gold response is given as an exemplar.
Lower PPL$_{retrieve}$ indicates that the exemplar-based generative model well leverages the provided exemplar to predict the gold response.
Dist-$n$~\cite{li2016diversity} is the ratio of distinct $n$-grams to a total number of $n$-grams for all the generated responses, which measures the degree of the diversity of the generated responses.
BLEU is adopted to measure the degree of the token overlap between the provided exemplar and the generated response pair ($z$, $r$).
A higher BLEU score indicates that the generator copies more from the provided exemplar while generating the response.

\subsection{Implementation Details}
We provide the details of our implementation in the Supplementary Material.
We released the source codes\footnote{\url{https://github.com/hyperconnect/corge}} of CORGE for the reproducibility of the conducted experiments.
\section{Experimental Results}\label{sec:6_experimental_results}
\subsection{Pair-wise Comparison Results}\label{subsec:pairwise_comparison}
Table~\ref{tab:2_pairwise} shows the pair-wise comparison results through the human evaluation.
When \textit{RetNRef} and \textit{MatToGen} adopt our proposed CORGE as their training method, they outperform all baselines except for a case of \textit{RetNRef}+\textit{CORGE} vs. \textit{KIF} on the informativeness.
In detail, \textit{RetNRef}+\textit{CORGE} and \textit{MatToGen}+\textit{CORGE} show better performance than \textit{RetNRef$_\alpha$} and \textit{MatToGen}, respectively, in both metrics. 
Especially, \textit{MatToGen}+\textit{CORGE} outperforms \textit{Bi-encoder 256M} and exceeds \textit{Blender 90M}, while \textit{MatToGen} performs worse than \textit{Bi-encoder 256M} and \textit{Blender 90M}.
Furthermore, CORGE enlarges the win rate of \textit{RetNRef$_\alpha$} for \textit{Blender 90M}.
These evaluation results demonstrate that CORGE leads the existing exemplar-based generative models to produce more fluent and informative responses.

\begin{table*}[t]
\centering
\footnotesize
\begin{tabular}{lcccccc}
\toprule
\textbf{Models} & \textbf{PPL}$\mathbf{_{gold}}$ & \textbf{PPL}$\mathbf{_{retrieve}}$ & \textbf{Dist-2} & \textbf{Dist-3} & \textbf{BLEU-2} & \textbf{BLEU-3}\\
\midrule
Blender 90M & 13.79 & 13.79 & 0.236 & 0.372 & - & -\\
Bi-encoder 256M & - &- & 0.681 & 0.881 & - & -\\
\midrule
\midrule
RetNRef & 8.518 & 13.37 & 0.256 & 0.386 & 0.030 & 0.009\\
RetNRef$_\alpha$ & 3.061 & 16.99 & 0.530 & 0.778 & 0.319 & 0.201\\
RetNRef + CORGE & 4.863 & 11.53 & 0.349 & 0.520 & 0.102 & 0.048\\
\midrule
MatToGen & 5.291 & 17.71 & 0.362 & 0.567 & 0.169 & 0.095\\
MatToGen + CORGE & 5.651 & 13.45 & 0.313 & 0.474 & 0.069 & 0.028 \\
\midrule
\midrule
RAG & 11.84 & 14.91 & 0.257 & 0.390 & 0.015 & 0.003\\
KIF & 12.11 & 15.18 & 0.238 & 0.363 & 0.002 & 0.000\\
\bottomrule
\end{tabular}
\caption{Automatic evaluation results. Since \textit{Blender 90M} can not utilize the exemplar, we report PPL calculated from $P_\mathbfcal{G}(r|c)$ in the place of PPL$_{gold}$ and PPL$_{retrieve}$.}
\label{tab:3_automatic_metrics}
\vspace*{-1em}
\end{table*}
\subsection{Investigating the Exemplar-based Generative Models with Automatic Metrics}
Through the automatic evaluation, we verify that existing exemplar-based generative models ignore the provided exemplar or generate responses over-fitted to the provided exemplar.
As shown in Table~\ref{tab:3_automatic_metrics}, \textit{RetNRef}+\textit{CORGE} and \textit{MatToGen}+\textit{CORGE} show lower PPL$_{retrieve}$ than \textit{Blender 90M}, which means that the exemplar-based generative models trained with CORGE make a better prediction of the gold response than \textit{Blender 90M} by utilizing the provided exemplar.
\textit{RetNRef}+\textit{CORGE} has a smaller degree of PPL$_{gold}$ and PPL$_{retrieve}$ than those of \textit{RetNRef}, which infers \textit{RetNRef}+\textit{CORGE} leverages the provided exemplar better than \textit{RetNRef}.
\textit{RetNRef$_\alpha$} has lower PPL$_{gold}$ than \textit{RetNRef}+\textit{CORGE}, however, \textit{RetNRef$_\alpha$} has higher PPL$_{retrieve}$ than \textit{RetNRef}+\textit{CORGE}.
This result demonstrates that \textit{RetNRef$_\alpha$} does not make good use of the retrieved exemplar except when the gold response is given as the retrieved exemplar.
From this observation, we claim that \textit{RetNRef$_\alpha$} generates a response highly over-fitted to the selected exemplar, which is caused by utilizing the gold response as an exemplar in the training phase.
The same goes for \textit{MatToGen}, where applying CORGE mitigates the over-fitting issue.

Higher Dist-$n$ of \textit{RetNRef}+\textit{CORGE} and \textit{MatToGen}+\textit{CORGE} compared to \textit{Blender 90M} shows that our exemplar-based generative models produce more diverse responses than the vanilla generative model.
Moreover, \textit{RetNRef}+\textit{CORGE} has higher Dist-$n$ than \textit{RetNRef}, which shows that utilizing the exemplars helps the generator diversify the responses.
Although \textit{RetNRef$_\alpha$} is the only one that achieves comparable Dist-$n$ to that of the vanilla retrieval model, \textit{Bi-encoder 256M}, it is derived from an over-fitting to the exemplar considering the gap between PPL$_{gold}$ and PPL$_{retrieve}$, resulting in the degradation of appropriateness and informativeness in human evaluation.

Average BLEU scores implicitly measure the overlap between the retrieved exemplar and the generated response; thus, a higher degree of BLEU indicates that the generator depends more on the retrieved exemplar.
\textit{RetNRef} shows a negligible BLEU score, which reaffirms that the model is almost not utilizing the retrieved exemplar.
\textit{RetNRef$_\alpha$} and \textit{MatToGen} have higher BLEU scores compared to \textit{RetNRef}+\textit{CORGE} and \textit{MatToGen}+\textit{CORGE}, respectively, which verifies that the former depends more on the retrieved exemplar than the latter.

\begin{table*}[ht]
\centering
\footnotesize
\setlength\tabcolsep{2.5pt}
\begin{tabular}{cccc}
\toprule
\multicolumn{1}{l}{\textbf{Input Context}} & \multicolumn{3}{l}{\begin{tabular}[l]{l}A: No way i do it a lot, at least when i am feeling well enough.\\B: Awesome. That’s the only thing i’ll never do. I do like the outdoors though.\\A: I’m a thrill seaker. Do you like extreme sports?\end{tabular}} \\ 
\midrule
\multicolumn{1}{l}{Bi-encoder 256M} & \multicolumn{3}{l}{I love extreme sports. Its the only real way I can get the adrenaline boost I need.} \\
\multicolumn{1}{l}{Blender 90M} & \multicolumn{3}{l}{No, I don 't think I' ve ever done that. I' d love to though.} \\
\midrule
\multicolumn{1}{l}{\textbf{Exemplar}} & \multicolumn{3}{l}{I love extreme sports. Its the only real way I can get the adrenaline boost I need.} \\
\multicolumn{1}{l}{RetNRef} & \multicolumn{3}{l}{I don' t really have a lot of time for sports. I spend most of my free time working.} \\
\midrule
\multicolumn{1}{l}{\textbf{Exemplar}} & \multicolumn{3}{l}{I love extreme sports. Its the only real way I can get the adrenaline boost I need.} \\
\multicolumn{1}{l}{RetNRef$\mathbf{_\alpha}$} & \multicolumn{3}{l}{I love extreme sports I need the adrenaline boost. Its the only way I can get the real adrenaline.} \\
\midrule
\multicolumn{1}{l}{\textbf{Exemplar}} & \multicolumn{3}{l}{I love extreme sports. Its the only real way I can get the adrenaline boost I need.} \\
\multicolumn{1}{l}{RetNRef + CORGE} & \multicolumn{3}{l}{I love extreme sports as well. It' s a great way to get out and enjoy life.} \\
\midrule
\multicolumn{1}{l}{\textbf{Exemplar}} & \multicolumn{3}{l}{nicde.} \\
\multicolumn{1}{l}{KIF} & \multicolumn{3}{l}{I don' t really have a lot of time for sports. I spend most of my time working.} \\
\midrule
\multicolumn{1}{l}{\textbf{Exemplar}} & \multicolumn{3}{l}{Do you like sports?} \\
\multicolumn{1}{l}{RAG} & \multicolumn{3}{l}{I don' t really have a lot of time for sports. I spend most of my time working.} \\
\bottomrule
\end{tabular}
\caption{Examples for model responses and provided exemplars corresponding to given input contexts.}
\label{tab:4_response_example}
\end{table*}

\subsection{Incapability of Knowledge-grounded Generative Models in Open-domain Conversation}\label{subsec:incap_knowledge_grounded}
The automatic evaluation results in Table~\ref{tab:3_automatic_metrics} confirm that knowledge-grounded generative models are ignoring the exemplar.
PPL$_{gold}$, PPL$_{retrieve}$, and Dist-$n$ of \textit{RAG} and \textit{KIF} have a similar degree to those of \textit{Blender 90M}, which implies that the exemplars are not providing useful information while generating the response. 
The average BLEU score also has a poor degree, indicating almost no overlap between the retrieved exemplars and the generated responses.
We explain that these results are originated from the difference between the open-domain conversation and knowledge-grounded generation tasks.
While training knowledge-grounded generative models, they use $P_\mathbfcal{R}(z, c)$ to fetch the external knowledge.
However, the generator also ignores the retrieved exemplar due to the one-to-many nature of the open-domain conversation.

\begin{figure}[t]
\centering\begin{tikzpicture}
\usepgfplotslibrary{colorbrewer}
\pgfplotsset{every axis/.append style={
                xlabel style={font=\small, yshift=1.0ex},
                ylabel style={font=\small, yshift=-1.0ex},
                ticklabel style={font=\footnotesize},
            }}
\begin{axis}[
    xlabel=Train steps,
    ylabel=Std. of Relevance Scores,
	height=4cm,
	width=8cm,
	legend style={nodes={scale=0.8}},
	legend cell align={left},    
	xmin=0,xmax=200,
	minor tick num=5,
    grid=both,
    grid style={line width=.1pt, draw=gray!10},
    major grid style={line width=.2pt,draw=gray!50},
    legend style={at={(0.95,0.7)},anchor=north east}
]
\addplot+[thick, mark=none] coordinates {
(1, 0.30787286162376404)
(2, 0.29945051670074463)
(3, 0.29462334513664246)
(4, 0.2984964847564697)
(5, 0.3075903058052063)
(6, 0.29605481028556824)
(7, 0.3235792815685272)
(8, 0.28689044713974)
(9, 0.2974361181259155)
(10, 0.3120894432067871)
(11, 0.31259092688560486)
(12, 0.3044970631599426)
(13, 0.2998543977737427)
(14, 0.30148079991340637)
(15, 0.3053538203239441)
(16, 0.2956791818141937)
(17, 0.3026978075504303)
(18, 0.30522990226745605)
(19, 0.30710628628730774)
(20, 0.31227147579193115)
(21, 0.29857712984085083)
(22, 0.3012184798717499)
(23, 0.30239182710647583)
(24, 0.30580446124076843)
(25, 0.2948966920375824)
(26, 0.2979905605316162)
(27, 0.3145357370376587)
(28, 0.2991662919521332)
(29, 0.30788254737854004)
(30, 0.31434935331344604)
(31, 0.3069985806941986)
(32, 0.29527726769447327)
(33, 0.3010154962539673)
(34, 0.31099873781204224)
(35, 0.2995854616165161)
(36, 0.3038436472415924)
(37, 0.298733115196228)
(38, 0.3004605770111084)
(39, 0.30403319001197815)
(40, 0.3025663197040558)
(41, 0.31238871812820435)
(42, 0.29774609208106995)
(43, 0.29373759031295776)
(44, 0.2992950975894928)
(45, 0.3032521903514862)
(46, 0.3063582479953766)
(47, 0.3113355040550232)
(48, 0.300505667924881)
(49, 0.30098554491996765)
(50, 0.3045964241027832)
(51, 0.3010490834712982)
(52, 0.2991517186164856)
(53, 0.30299142003059387)
(54, 0.3023557960987091)
(55, 0.2912455201148987)
(56, 0.3048962354660034)
(57, 0.3053196966648102)
(58, 0.30716925859451294)
(59, 0.30089735984802246)
(60, 0.3101823329925537)
(61, 0.3125477731227875)
(62, 0.29829081892967224)
(63, 0.29555100202560425)
(64, 0.2999352514743805)
(65, 0.30183011293411255)
(66, 0.3006073236465454)
(67, 0.30248838663101196)
(68, 0.30678603053092957)
(69, 0.3007097542285919)
(70, 0.30257123708724976)
(71, 0.3052942454814911)
(72, 0.2965294420719147)
(73, 0.3082159459590912)
(74, 0.3038490116596222)
(75, 0.3089582920074463)
(76, 0.2992233633995056)
(77, 0.2966049015522003)
(78, 0.3001835346221924)
(79, 0.30530470609664917)
(80, 0.2937258183956146)
(81, 0.30000442266464233)
(82, 0.3082583546638489)
(83, 0.29842472076416016)
(84, 0.3011534512042999)
(85, 0.299732506275177)
(86, 0.31117546558380127)
(87, 0.30242919921875)
(88, 0.3012065589427948)
(89, 0.31200703978538513)
(90, 0.30902209877967834)
(91, 0.3035839796066284)
(92, 0.3157944083213806)
(93, 0.2982501983642578)
(94, 0.30516159534454346)
(95, 0.3033175766468048)
(96, 0.29049763083457947)
(97, 0.2976120412349701)
(98, 0.3000328838825226)
(99, 0.307137668132782)
(100, 0.3021675944328308)
(101, 0.3039955794811249)
(102, 0.3028031587600708)
(103, 0.29719218611717224)
(104, 0.30086421966552734)
(105, 0.2864370048046112)
(106, 0.29630255699157715)
(107, 0.2943952679634094)
(108, 0.3041090965270996)
(109, 0.30795755982398987)
(110, 0.30018043518066406)
(111, 0.29929429292678833)
(112, 0.30056825280189514)
(113, 0.3154616057872772)
(114, 0.30400151014328003)
(115, 0.29783520102500916)
(116, 0.3100566864013672)
(117, 0.3021124601364136)
(118, 0.2966783940792084)
(119, 0.3048628568649292)
(120, 0.29435107111930847)
(121, 0.30275338888168335)
(122, 0.3059922456741333)
(123, 0.2965496778488159)
(124, 0.29531440138816833)
(125, 0.30945372581481934)
(126, 0.3016504943370819)
(127, 0.30077993869781494)
(128, 0.2987816333770752)
(129, 0.31163540482521057)
(130, 0.30681777000427246)
(131, 0.2908846139907837)
(132, 0.3144809305667877)
(133, 0.2900582551956177)
(134, 0.30509424209594727)
(135, 0.29739242792129517)
(136, 0.2919697165489197)
(137, 0.3058396577835083)
(138, 0.29587528109550476)
(139, 0.29472121596336365)
(140, 0.29771339893341064)
(141, 0.292427122592926)
(142, 0.3056946098804474)
(143, 0.2962675988674164)
(144, 0.30423104763031006)
(145, 0.3018724024295807)
(146, 0.29584047198295593)
(147, 0.30429646372795105)
(148, 0.30795833468437195)
(149, 0.3067956566810608)
(150, 0.3010362386703491)
(151, 0.297851026058197)
(152, 0.3098994791507721)
(153, 0.30752304196357727)
(154, 0.30770665407180786)
(155, 0.31200194358825684)
(156, 0.30860865116119385)
(157, 0.30740100145339966)
(158, 0.3085331618785858)
(159, 0.3081599771976471)
(160, 0.3029053807258606)
(161, 0.30059614777565)
(162, 0.3056573271751404)
(163, 0.29592615365982056)
(164, 0.3026343882083893)
(165, 0.2958662807941437)
(166, 0.2899484932422638)
(167, 0.3068487048149109)
(168, 0.29947543144226074)
(169, 0.30657288432121277)
(170, 0.3132902979850769)
(171, 0.30270442366600037)
(172, 0.29667928814888)
(173, 0.30363765358924866)
(174, 0.30495455861091614)
(175, 0.29805806279182434)
(176, 0.2981577515602112)
(177, 0.3066595792770386)
(178, 0.2996821105480194)
(179, 0.30234062671661377)
(180, 0.29149606823921204)
(181, 0.30179429054260254)
(182, 0.2997599244117737)
(183, 0.3152892589569092)
(184, 0.31231677532196045)
(185, 0.3012559413909912)
(186, 0.292756050825119)
(187, 0.29436808824539185)
(188, 0.30176836252212524)
(189, 0.30562666058540344)
(190, 0.30120041966438293)
(191, 0.3012945055961609)
(192, 0.297085702419281)
(193, 0.3077244460582733)
(194, 0.3034518361091614)
(195, 0.30468112230300903)
(196, 0.30361950397491455)
(197, 0.3058689534664154)
(198, 0.30297690629959106)
(199, 0.3053900897502899)
(200, 0.30048811435699463)
};
\addplot+[thick, mark=none ] coordinates {
(1, 0.2930320203304291)
(2, 0.299173504114151)
(3, 0.29581838846206665)
(4, 0.29929080605506897)
(5, 0.30089932680130005)
(6, 0.2959502935409546)
(7, 0.2921333312988281)
(8, 0.2857440114021301)
(9, 0.31116580963134766)
(10, 0.30037787556648254)
(11, 0.2957988679409027)
(12, 0.3013654351234436)
(13, 0.3026842474937439)
(14, 0.2959003150463104)
(15, 0.30327409505844116)
(16, 0.28716349601745605)
(17, 0.29141783714294434)
(18, 0.2997259795665741)
(19, 0.2901099920272827)
(20, 0.28446701169013977)
(21, 0.27879953384399414)
(22, 0.28067082166671753)
(23, 0.2771639823913574)
(24, 0.26717114448547363)
(25, 0.2689744532108307)
(26, 0.25159379839897156)
(27, 0.2546791732311249)
(28, 0.24678684771060944)
(29, 0.2436869740486145)
(30, 0.25772324204444885)
(31, 0.2371242791414261)
(32, 0.23505547642707825)
(33, 0.24061186611652374)
(34, 0.23691213130950928)
(35, 0.2418433576822281)
(36, 0.23724612593650818)
(37, 0.24164022505283356)
(38, 0.23081299662590027)
(39, 0.23612767457962036)
(40, 0.23011071979999542)
(41, 0.22487877309322357)
(42, 0.22220660746097565)
(43, 0.22251982986927032)
(44, 0.22732526063919067)
(45, 0.22540079057216644)
(46, 0.2291068285703659)
(47, 0.2258814424276352)
(48, 0.22695453464984894)
(49, 0.2229345291852951)
(50, 0.23119230568408966)
(51, 0.2282065898180008)
(52, 0.22128669917583466)
(53, 0.2400224655866623)
(54, 0.23207110166549683)
(55, 0.22514140605926514)
(56, 0.2238778918981552)
(57, 0.21725288033485413)
(58, 0.2251332402229309)
(59, 0.21953506767749786)
(60, 0.22451342642307281)
(61, 0.23044301569461823)
(62, 0.2320326715707779)
(63, 0.2402094304561615)
(64, 0.23191943764686584)
(65, 0.2425876259803772)
(66, 0.21778257191181183)
(67, 0.21238714456558228)
(68, 0.2279706746339798)
(69, 0.2066701054573059)
(70, 0.2184167057275772)
(71, 0.21929727494716644)
(72, 0.23636803030967712)
(73, 0.21934019029140472)
(74, 0.22432205080986023)
(75, 0.23869481682777405)
(76, 0.22954337298870087)
(77, 0.2374301552772522)
(78, 0.22349587082862854)
(79, 0.2382003515958786)
(80, 0.22692210972309113)
(81, 0.23285256326198578)
(82, 0.23533973097801208)
(83, 0.2146786004304886)
(84, 0.2179345041513443)
(85, 0.22235107421875)
(86, 0.21837905049324036)
(87, 0.22186027467250824)
(88, 0.22585225105285645)
(89, 0.24232345819473267)
(90, 0.23864957690238953)
(91, 0.23454132676124573)
(92, 0.2179335206747055)
(93, 0.21763984858989716)
(94, 0.22597423195838928)
(95, 0.24816958606243134)
(96, 0.22034765779972076)
(97, 0.23548197746276855)
(98, 0.23230288922786713)
(99, 0.2196437418460846)
(100, 0.23435042798519135)
(101, 0.23793916404247284)
(102, 0.23998410999774933)
(103, 0.2224268615245819)
(104, 0.22938470542430878)
(105, 0.23241205513477325)
(106, 0.22296009957790375)
(107, 0.2386094480752945)
(108, 0.22149847447872162)
(109, 0.23880092799663544)
(110, 0.22062087059020996)
(111, 0.2401147186756134)
(112, 0.2388097196817398)
(113, 0.2300032079219818)
(114, 0.22657287120819092)
(115, 0.24859662353992462)
(116, 0.23346063494682312)
(117, 0.22817189991474152)
(118, 0.221185564994812)
(119, 0.23931743204593658)
(120, 0.238593190908432)
(121, 0.22457349300384521)
(122, 0.23468373715877533)
(123, 0.21754267811775208)
(124, 0.2464069128036499)
(125, 0.2353123128414154)
(126, 0.23264199495315552)
(127, 0.2266143262386322)
(128, 0.2286633998155594)
(129, 0.2185850739479065)
(130, 0.23513679206371307)
(131, 0.21916769444942474)
(132, 0.2357749044895172)
(133, 0.232798770070076)
(134, 0.23066288232803345)
(135, 0.22876809537410736)
(136, 0.24061031639575958)
(137, 0.24004973471164703)
(138, 0.2342524379491806)
(139, 0.2275519073009491)
(140, 0.23666468262672424)
(141, 0.24237900972366333)
(142, 0.2378953993320465)
(143, 0.23266634345054626)
(144, 0.23410652577877045)
(145, 0.22118519246578217)
(146, 0.23329883813858032)
(147, 0.2397492229938507)
(148, 0.24412617087364197)
(149, 0.2383432686328888)
(150, 0.22222641110420227)
(151, 0.22986522316932678)
(152, 0.2327292561531067)
(153, 0.2294214516878128)
(154, 0.23203225433826447)
(155, 0.233732208609581)
(156, 0.22971715033054352)
(157, 0.2492814064025879)
(158, 0.22676753997802734)
(159, 0.23627564311027527)
(160, 0.2307075560092926)
(161, 0.22794108092784882)
(162, 0.220607191324234)
(163, 0.22192512452602386)
(164, 0.23553451895713806)
(165, 0.23555922508239746)
(166, 0.22561004757881165)
(167, 0.22692765295505524)
(168, 0.2391153872013092)
(169, 0.23253637552261353)
(170, 0.2175806611776352)
(171, 0.23332130908966064)
(172, 0.2214507758617401)
(173, 0.21793852746486664)
(174, 0.23760372400283813)
(175, 0.2458697110414505)
(176, 0.24573053419589996)
(177, 0.22798357903957367)
(178, 0.23103544116020203)
(179, 0.23218871653079987)
(180, 0.23048731684684753)
(181, 0.2278224527835846)
(182, 0.21933214366436005)
(183, 0.23473455011844635)
(184, 0.23270860314369202)
(185, 0.23886865377426147)
(186, 0.23036807775497437)
(187, 0.23397491872310638)
(188, 0.2209545224905014)
(189, 0.23999565839767456)
(190, 0.2308792769908905)
(191, 0.22987757623195648)
(192, 0.23772557079792023)
(193, 0.2281276285648346)
(194, 0.23611056804656982)
(195, 0.23165197670459747)
(196, 0.24752622842788696)
(197, 0.25136759877204895)
(198, 0.24555692076683044)
(199, 0.2538634240627289)
(200, 0.23433977365493774)
};
\addplot+[thick, mark=none] coordinates {
(1, 0.23396039009094238)
(2, 0.25009000301361084)
(3, 0.2633112967014313)
(4, 0.25651562213897705)
(5, 0.25713539123535156)
(6, 0.2659932076931)
(7, 0.26942840218544006)
(8, 0.25658345222473145)
(9, 0.24987225234508514)
(10, 0.24964012205600739)
(11, 0.2523118555545807)
(12, 0.26053860783576965)
(13, 0.24815651774406433)
(14, 0.24878911674022675)
(15, 0.24229927361011505)
(16, 0.2498779147863388)
(17, 0.24618293344974518)
(18, 0.24996858835220337)
(19, 0.23817653954029083)
(20, 0.2315824031829834)
(21, 0.2262626737356186)
(22, 0.22075749933719635)
(23, 0.21547527611255646)
(24, 0.2273392528295517)
(25, 0.22005203366279602)
(26, 0.20980481803417206)
(27, 0.2125902771949768)
(28, 0.2105138748884201)
(29, 0.20827414095401764)
(30, 0.19725441932678223)
(31, 0.20426379144191742)
(32, 0.2014286369085312)
(33, 0.185996875166893)
(34, 0.1886865645647049)
(35, 0.18807843327522278)
(36, 0.1894819438457489)
(37, 0.16849389672279358)
(38, 0.1759672909975052)
(39, 0.18232221901416779)
(40, 0.1677934229373932)
(41, 0.16633321344852448)
(42, 0.16955478489398956)
(43, 0.1677960604429245)
(44, 0.1660267859697342)
(45, 0.16281896829605103)
(46, 0.15099819004535675)
(47, 0.14797495305538177)
(48, 0.15319812297821045)
(49, 0.13916534185409546)
(50, 0.15163953602313995)
(51, 0.14916686713695526)
(52, 0.14455166459083557)
(53, 0.13540127873420715)
(54, 0.1374409943819046)
(55, 0.13576740026474)
(56, 0.12399900704622269)
(57, 0.11701701581478119)
(58, 0.12143613398075104)
(59, 0.12407936155796051)
(60, 0.11049138754606247)
(61, 0.11368585377931595)
(62, 0.11638939380645752)
(63, 0.11842487007379532)
(64, 0.10743391513824463)
(65, 0.10803607851266861)
(66, 0.1118980273604393)
(67, 0.10269944369792938)
(68, 0.1113307997584343)
(69, 0.11220517754554749)
(70, 0.1124596819281578)
(71, 0.10754309594631195)
(72, 0.09794813394546509)
(73, 0.10048513114452362)
(74, 0.0989660918712616)
(75, 0.10246152430772781)
(76, 0.10106270015239716)
(77, 0.10355991125106812)
(78, 0.10258728265762329)
(79, 0.10088522732257843)
(80, 0.09806562960147858)
(81, 0.09438793361186981)
(82, 0.09652019292116165)
(83, 0.09275322407484055)
(84, 0.09363614022731781)
(85, 0.09791900217533112)
(86, 0.0930882915854454)
(87, 0.09261037409305573)
(88, 0.09071139246225357)
(89, 0.09547948092222214)
(90, 0.08995188772678375)
(91, 0.09237363189458847)
(92, 0.0965399220585823)
(93, 0.08768516033887863)
(94, 0.09084996581077576)
(95, 0.08873007446527481)
(96, 0.08624886721372604)
(97, 0.08667482435703278)
(98, 0.08496324717998505)
(99, 0.0838913545012474)
(100, 0.0820772722363472)
(101, 0.08294395357370377)
(102, 0.08683692663908005)
(103, 0.08740738034248352)
(104, 0.08188901096582413)
(105, 0.08758167177438736)
(106, 0.09464409202337265)
(107, 0.08260266482830048)
(108, 0.09058792144060135)
(109, 0.07012232393026352)
(110, 0.07194489985704422)
(111, 0.0724492222070694)
(112, 0.07025177031755447)
(113, 0.06657887250185013)
(114, 0.07027412950992584)
(115, 0.07346190512180328)
(116, 0.07551680505275726)
(117, 0.07625561207532883)
(118, 0.07215127348899841)
(119, 0.08450712263584137)
(120, 0.08454644680023193)
(121, 0.07938947528600693)
(122, 0.0833878144621849)
(123, 0.08774513006210327)
(124, 0.08993551880121231)
(125, 0.07770932465791702)
(126, 0.08286652714014053)
(127, 0.07553530484437943)
(128, 0.08430910855531693)
(129, 0.07570338994264603)
(130, 0.0737888365983963)
(131, 0.07818231731653214)
(132, 0.073881134390831)
(133, 0.0742482990026474)
(134, 0.07619117200374603)
(135, 0.08033652603626251)
(136, 0.08438732475042343)
(137, 0.0838804617524147)
(138, 0.08389833569526672)
(139, 0.07853438705205917)
(140, 0.07408285140991211)
(141, 0.07225263118743896)
(142, 0.0708606168627739)
(143, 0.07511834800243378)
(144, 0.06827764958143234)
(145, 0.07748252898454666)
(146, 0.08279509842395782)
(147, 0.0876317024230957)
(148, 0.08570259809494019)
(149, 0.08346720039844513)
(150, 0.08422534167766571)
(151, 0.08155696839094162)
(152, 0.08252549916505814)
(153, 0.08144202828407288)
(154, 0.07931715250015259)
(155, 0.07569455355405807)
(156, 0.0749114379286766)
(157, 0.0812515839934349)
(158, 0.07501036673784256)
(159, 0.07649649679660797)
(160, 0.08124230057001114)
(161, 0.08048111945390701)
(162, 0.08541730046272278)
(163, 0.08203042298555374)
(164, 0.08040672540664673)
(165, 0.08836372941732407)
(166, 0.07863261550664902)
(167, 0.07557117193937302)
(168, 0.0786033496260643)
(169, 0.08456185460090637)
(170, 0.08193746209144592)
(171, 0.07945811748504639)
(172, 0.08704975247383118)
(173, 0.0871778130531311)
(174, 0.08785068243741989)
(175, 0.09077004343271255)
(176, 0.08853723108768463)
(177, 0.07819990813732147)
(178, 0.08310474455356598)
(179, 0.07615268230438232)
(180, 0.07577649503946304)
(181, 0.06878924369812012)
(182, 0.07213904708623886)
(183, 0.0784476101398468)
(184, 0.07757291942834854)
(185, 0.08166975528001785)
(186, 0.08672593533992767)
(187, 0.08613621443510056)
(188, 0.08202970027923584)
(189, 0.08303189277648926)
(190, 0.08031578361988068)
(191, 0.08178143203258514)
(192, 0.0822061225771904)
(193, 0.09049151092767715)
(194, 0.08908352255821228)
(195, 0.07672429829835892)
(196, 0.08223381638526917)
(197, 0.08008911460638046)
(198, 0.07185041904449463)
(199, 0.06823214888572693)
(200, 0.07345394790172577)
};
\legend{\textit{Ours}, \textit{Ours + joint}, \textit{RAG}}
\end{axis}
\end{tikzpicture}
\vspace*{-1em}
\caption{The standard deviation of the normalized retriever score gets smaller when we jointly train the retriever for exemplar-based generative models.
\textit{Ours} stands for \textit{RetNRef}+\textit{CORGE}, and \textit{joint} indicates jointly training the retriever with the generator.}
\vspace*{-1em}
\label{fig:4_retriever_flat}
\end{figure}
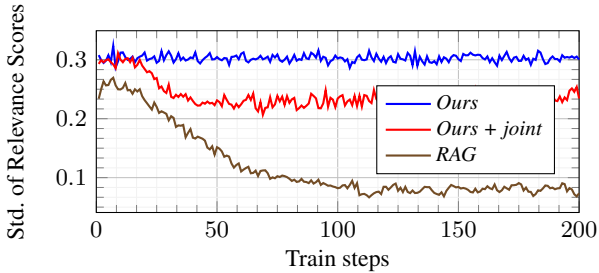
\begin{table*}[t]
\centering
\footnotesize
\begin{tabular}{lccccccc}
\toprule
\textbf{Models} & \textbf{PPL}$\mathbf{_{gold}}$ & \textbf{PPL}$\mathbf{_{retrieve}}$  & \textbf{Dist-2} & \textbf{Dist-3} & \textbf{BLEU-2} & \textbf{BLEU-3}\\
\midrule
RetNRef + CORGE & 4.863 & 11.53 & 0.349 & 0.520 & 0.102 & 0.048\\
RetNRef + CORGE $-$ RS & 6.482 & 11.75 & 0.316 & 0.478 & 0.074 & 0.031\\
RetNRef + CORGE $-$ kNE & 8.657 & 13.82 & 0.250 & 0.380 & 0.034 & 0.010\\
RetNRef + CORGE $-$ JF & 1.698 & 32.91 & 0.537 & 0.785 & 0.332 & 0.207\\
\bottomrule
\end{tabular}
\caption{Results of the ablation study. $-$RS, $-$kNE, and $-$JF denote that relevance score (RS), $k$NE, and Jaccard filter (JF) are removed from CORGE, respectively.}
\label{tab:5_ablation_automatic_metrics}
\vspace*{-1.5em}
\end{table*}
In addition, we observe that jointly training the retriever with the generator causes the retriever stuck in the local minima.
As shown in Figure~\ref{fig:4_retriever_flat}, the standard deviation of normalized relevance scores $P_\mathbfcal{R}(z,c)$ computed by the retriever almost gets near zero when the retriever of RAG is jointly trained.
A smaller standard deviation means the relevance scores are getting flattened.
Although knowledge-grounded generative models empirically have shown that jointly training the retriever and generator improves the performance in knowledge-intensive NLP tasks~\cite{NEURIPS2020_6b493230}, in open-domain conversation, the retrieved exemplars are ignored.
Thus, the retriever learns to produce an uninformative relevance score.
As a result, the retriever collapses, which means the retriever may return inappropriate exemplars to the generator (also shown in the example of KIF and RAG in Table~\ref{tab:4_response_example}).
Intriguingly, jointly training the retriever with CORGE also causes the retriever scores to be flattened, as shown in Figure~\ref{fig:4_retriever_flat}, and we empirically observe the minor collapse of the retriever as we experienced in RAG as well.
Thus, CORGE does not jointly train the retriever.

\subsection{Ablation Study}
To verify the effectiveness of each component in CORGE, we conduct the ablation study.
In Table~\ref{tab:5_ablation_automatic_metrics}, PPL$_{retrieve}$ from \textit{RetNRef}+\textit{CORGE} is lower than any other ablation counterparts, which confirms each component contributes to predicting the responses.
\textit{RetNRef}+\textit{CORGE}$-$RS and \textit{RetNRef}+\textit{CORGE}$-$\textit{k}NE have a higher degree of PPL$_{retrieve}$ and PPL$_{gold}$, which indicates \textit{RS} and \textit{kNE} help the generator to utilize the exemplar while generating the response.
\textit{RetNRef}+\textit{CORGE}$-$JF provides a strong signal of over-fitting, where it has extremely low PPL$_{gold}$ but exceptionally high PPL$_{retrieve}$.
Dist-$n$ shows our model produces the most diverse responses among the models except \textit{RetNRef}+\textit{CORGE}$-$JF, where \textit{RetNRef}+\textit{CORGE}$-$JF excessively copies the tokens from the retrieved exemplar.
The average BLEU scores also show the same trend, where reaffirms the effect of the components of CORGE.
\section{Conclusion}\label{sec:7_conclusion}
In this paper, we propose a generally applicable training method for exemplar-based generative models to alleviate their disadvantages derived from the one-to-many problem of the open-domain conversation.
Our proposed training method selects exemplars that are semantically relevant but lexically distanced from the gold response and weights those exemplars with the relevance score measured by the retriever.
Through the extensive analysis, including pair-wise human evaluation, we verify that our method improves the performance of existing exemplar-based generative models in terms of appropriateness and informativeness.

\bibliography{aaai22}
\clearpage
\newpage
\appendix
\section{Implementation Details}
\begin{figure*}[t]
\centering
\includegraphics[width=\textwidth]{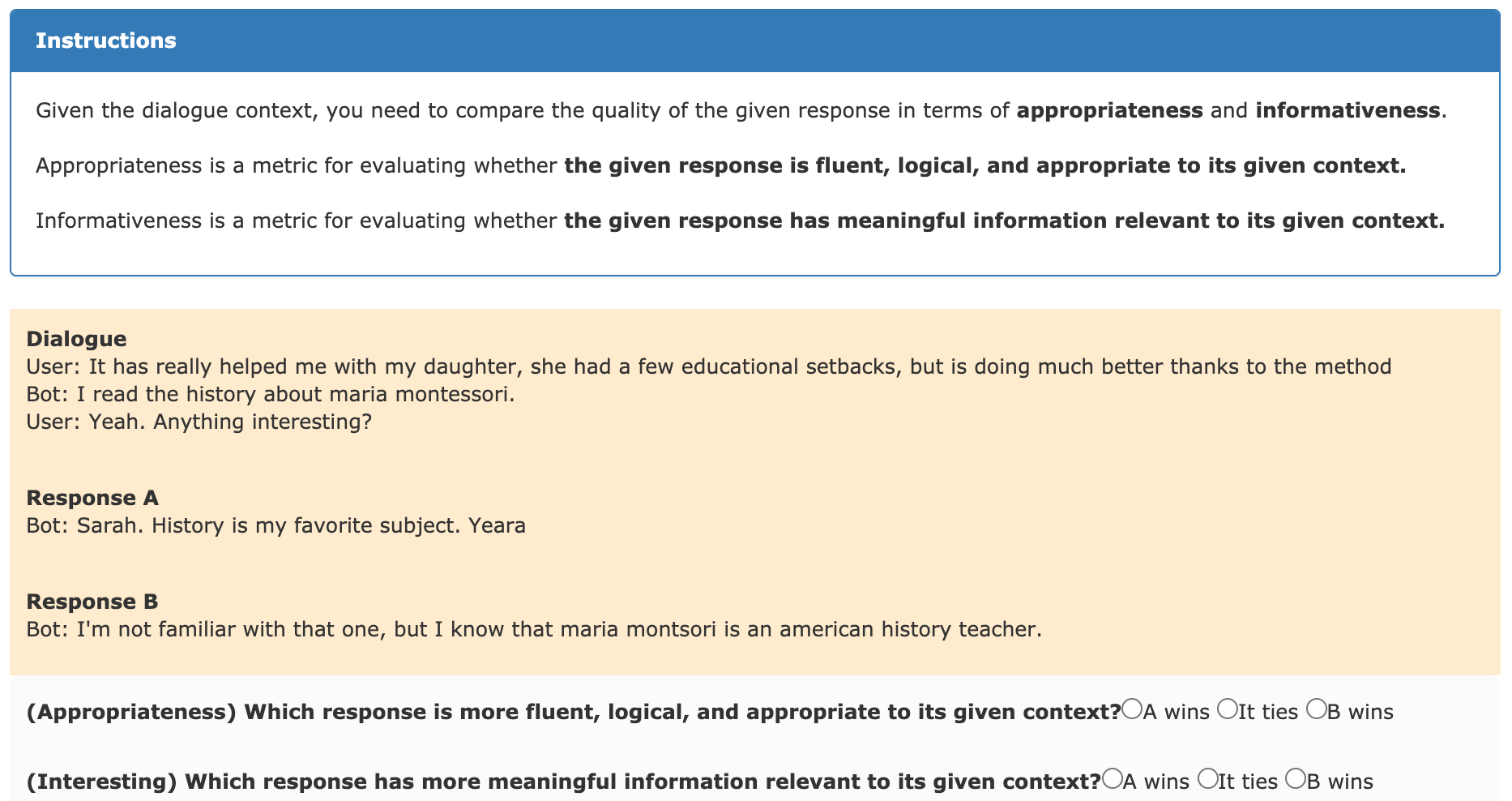}
\caption{The interface of pairwise human evaluation for appropriateness and informativeness.}
\label{fig:5_mturk}
\end{figure*}

\begin{table*}[ht]
\centering
\footnotesize
\setlength\tabcolsep{2.5pt}
\begin{tabular}{cccc}
\toprule
\multicolumn{1}{l}{\textbf{Input Context}} & \multicolumn{3}{l}{\begin{tabular}[l]{l}A: I crash around 230 to 300. I don't have much of a social life.\\ B: Neither do I, you are not alone\\A: I am alone but I'm not lonely. I have a dog who's great company.\end{tabular}} \\ 
\midrule
\multicolumn{1}{l}{Bi-encoder 256M} & \multicolumn{3}{l}{Any friends? My dog is one of mine.} \\
\multicolumn{1}{l}{Blender 90M} & \multicolumn{3}{l}{That 's good to hear, I 've a dog too, he' s my best friend.} \\
\midrule
\multicolumn{1}{l}{\textbf{Exemplar}} & \multicolumn{3}{l}{Any friends? My dog is one of mine.} \\
\multicolumn{1}{l}{RetNRef} & \multicolumn{3}{l}{I am sorry to hear that. Do you have any hobbies that you like to do?} \\
\multicolumn{1}{l}{RetNRef$\mathbf{_\alpha}$} & \multicolumn{3}{l}{My dog is one of my friends? Is yours one of any of his friends? lol.} \\
\multicolumn{1}{l}{RetNRef + CORGE} & \multicolumn{3}{l}{What kind of dog is it? I' ve got a chihuahua and she' s my best friend.} \\
\midrule

\multicolumn{1}{l}{\textbf{Exemplar}} & \multicolumn{3}{l}{I love dogs too. My best friend is my dog.} \\
\multicolumn{1}{l}{MatToGen} & \multicolumn{3}{l}{I love my best friend, he' s the best dog I've ever had. What is your best friend?} \\
\multicolumn{1}{l}{MatToGen + CORGE} & \multicolumn{3}{l}{I love dogs too, my best friend is my dog. Do you have any hobbies?} \\
\midrule
\multicolumn{1}{l}{\textbf{Exemplar}} & \multicolumn{3}{l}{nicde.} \\
\multicolumn{1}{l}{KIF} & \multicolumn{3}{l}{I am sorry to hear that. Do you have any hobbies that you like to do?} \\
\midrule
\multicolumn{1}{l}{\textbf{Exemplar}} & \multicolumn{3}{l}{I'm usually alone because I don't like my family.} \\
\multicolumn{1}{l}{RAG} & \multicolumn{3}{l}{I am sorry to hear that. Do you have any hobbies that you like to do?} \\
\bottomrule
\toprule
\multicolumn{1}{l}{\textbf{Input Context}} & \multicolumn{3}{l}{\begin{tabular}[l]{l}A: Sounds like a horror movie, my favorite kind\\B: Haha, yeah he melts my heart.\\A: Did you do anything fun for Halloween?\end{tabular}} \\ 
\midrule
\multicolumn{1}{l}{Bi-encoder 256M} & \multicolumn{3}{l}{I enjoy Halloween, it's fun to dress up. Do you like it?} \\
\multicolumn{1}{l}{Blender 90M} & \multicolumn{3}{l}{I went to a halloween party with my friends. It was a lot of fun. What about you?} \\
\midrule
\multicolumn{1}{l}{\textbf{Exemplar}} & \multicolumn{3}{l}{I enjoy Halloween, it's fun to dress up. Do you like it?} \\
\multicolumn{1}{l}{RetNRef} & \multicolumn{3}{l}{No, I don 't think I could do it. I' m not a big fan of halloween.} \\
\multicolumn{1}{l}{RetNRef$\mathbf{_\alpha}$} & \multicolumn{3}{l}{I like it, it' s fun to dress up. Do you like halloween? I enjoy dressing up.} \\
\multicolumn{1}{l}{RetNRef + CORGE} & \multicolumn{3}{l}{I like to dress up as a clown, but I don' t think I' ll be doing that this year.} \\
\midrule

\multicolumn{1}{l}{\textbf{Exemplar}} & \multicolumn{3}{l}{No we went to a haunted house and I got chased often.} \\
\multicolumn{1}{l}{MatToGen} & \multicolumn{3}{l}{No we never got to go to the haunted house we went to once and got chased by a ghost.} \\
\multicolumn{1}{l}{MatToGen + CORGE} & \multicolumn{3}{l}{No, I don 't like to go to haunted houses. I' m scared of them.} \\
\midrule
\multicolumn{1}{l}{\textbf{Exemplar}} & \multicolumn{3}{l}{nicde.} \\
\multicolumn{1}{l}{KIF} & \multicolumn{3}{l}{Yes, I went to a haunted house and watched all the horror movies. It was so scary!} \\
\midrule
\multicolumn{1}{l}{\textbf{Exemplar}} & \multicolumn{3}{l}{Are you doing anything cool for halloween} \\
\multicolumn{1}{l}{RAG} & \multicolumn{3}{l}{No, I don' t think I' m going to. I' ve never been a fan of halloween.} \\
\bottomrule
\toprule
\multicolumn{1}{l}{\textbf{Input Context}} & \multicolumn{3}{l}{\begin{tabular}[l]{l}A: Oh God no! I can't stay tied down to one woman. Why would you want to do that?\\B: I know right? Most people consider marriage to be involving 2 people but in certain parts of the \\world that varies between cultures and religions, so it doesn't seem
so bad\\A: Marriage is very good.\end{tabular}} \\ 
\midrule
\multicolumn{1}{l}{Bi-encoder 256M} & \multicolumn{3}{l}{I agree, but in some cultures, marriage is recommended and unmarried folks are frowned upon!} \\
\multicolumn{1}{l}{Blender 90M} & \multicolumn{3}{l}{I agree! I' m not sure if i could do it though. I don' t think I could.} \\
\midrule
\multicolumn{1}{l}{\textbf{Exemplar}} & \multicolumn{3}{l}{I agree, but in some cultures, marriage is recommended and unmarried folks are frowned upon!} \\
\multicolumn{1}{l}{RetNRef} & \multicolumn{3}{l}{I agree. I think it' s important for people to have a healthy relationship with each other.} \\
\multicolumn{1}{l}{RetNRef$\mathbf{_\alpha}$} & \multicolumn{3}{l}{I agree, but some folks are recommended to marry in some cultures, and some are frown upon!} \\
\multicolumn{1}{l}{RetNRef + CORGE} & \multicolumn{3}{l}{I agree! In some cultures marriage is recommended, but in some countries it' s not recommended.} \\
\midrule

\multicolumn{1}{l}{\textbf{Exemplar}} & \multicolumn{3}{l}{I agree. I think it' s important for people to have a healthy relationship with each other.} \\
\multicolumn{1}{l}{MatToGen} & \multicolumn{3}{l}{I agree, some cultures are more strict than others, but i agree that some religions are more lenient.} \\
\multicolumn{1}{l}{MatToGen + CORGE} & \multicolumn{3}{l}{I agree! In some cultures marriage is recommended by some religions, but I don' t believe in it!} \\
\midrule
\multicolumn{1}{l}{\textbf{Exemplar}} & \multicolumn{3}{l}{nicde.} \\
\multicolumn{1}{l}{KIF} & \multicolumn{3}{l}{I agree! I think it' s important to remember that marriage is a legal, social, and financial union.} \\
\midrule
\multicolumn{1}{l}{\textbf{Exemplar}} & \multicolumn{3}{l}{That is good and great} \\
\multicolumn{1}{l}{RAG} & \multicolumn{3}{l}{I agree! I think it' s important for people to have a healthy relationship with each other.} \\
\bottomrule
\end{tabular}
\caption{Additional examples for model responses and provided exemplars corresponding to given input contexts.}
\label{tab:6_additional_results}
\end{table*}
\subsection{How the Retriever Calculates the Scores}
Our retriever follows the architecture of Bi-encoder~\cite{mazare2018training}, and the score $S_\mathbfcal{R}(z, c)$ and $S_\mathbfcal{R'}(z, r)$ are calculated as follows:
\begin{equation}
\begin{split}
    S_\mathbfcal{R}(z, c) &= d(z) \cdot q(c), \\
    S_\mathbfcal{R'}(z, r) &= d(z) \cdot d(r), \\
    d(z) &= \mathbf{BERT}_r (z),\\
    d(r) &= \mathbf{BERT}_r (r), \\
    q(c) &= \mathbf{BERT}_c (c),
\end{split}
\end{equation}
where $d(z)$ and $d(r)$ are encoded vectors produced by response encoder $\mathbf{BERT}_r$ and $q(c)$ is an encoded vector produced by context encoder $\mathbf{BERT}_c$.
The notation $\mathbfcal{R'}$ indicates that it only uses the response encoder instead of using the context encoder together.
CORGE is not limited to use Bi-encoder as a retriever and can be applied to other types of a retriever (e.g. Poly-encoder~\cite{humeau2019poly}).

\subsection{Model Details}
As we mentioned in Section 5.2, we employ Bi-encoder 256M and Blender 90M as a retriever and a generator of each exemplar-based generative model, respectively.
For MatToGen, additional MLP layers are added to the retriever, as follows the details in~\citet{cai2019retrieval}.
When training the models, weights of the retriever and the generator are initialized with the pre-trained Bi-encoder 256M and Blender 90M, respectively,
For Blender 90M, we use the model released by ParlAI~\cite{miller2017parlai}, which is fine-tuned on the BST$+$ dataset.
For Bi-encoder 256M, we fine-tune the model released by ParlAI on the BST$+$ dataset, and we follow the hyperparameter settings of~\citet{humeau2019poly}, which are implemented in the ParlAI library.
The pre-defined response set is constructed from the BST$+$ training set, which contains about 400K responses.
We use NVIDIA DGX Station A100 for training the models.

\subsection{Hyperparameters}
When training exemplar-based generative models with CORGE, five ($k$=5) exemplars are utilized for each training instance.
The exemplar-based generators are trained with a batch size of 32 and an initial learning rate of 7e-6, and the learning rate is decayed in half when the training loss meets the plateau.
The model is trained until there is no progress in the validation PPL.

\subsection{Generation Strategy}
When we generate samples using generative model, exemplar-based generative models, and knowledge-grounded generative models, we adopt a beam decoding strategy which is widely used in generative models~\cite{graves2012sequence}.
Following~\cite{roller2021recipes}, we choose a minimum beam length and a beam size as 20 BPE tokens and 10, respectively, and use tri-gram beam blocking on context and response blocks.
During the inference phase, both exemplar-based generative models and knowledge-grounded generative models use the top-1 scoring candidate as an exemplar chosen from utilizing the relevance score $S_\mathbfcal{R}(z, c)$.

\section{Evaluation Details}
We prepare dialogue cases that have three-turn input contexts and the gold response from the BST and evaluate them by human pair-wise comparison and automatic evaluation.
There are 980 test cases, and we randomly choose 100 test cases for the human evaluation.

\subsection{Pair-wise Human Evaluation}
As we described in Section 5.3, we use Amazon Mechanical Turk to collect the annotations.
Each test case is rated by three annotators to improve the robustness of the evaluation result.
We set a maximum number of annotations per worker in order to reduce the potential bias.
To control the quality of the annotations, we only allowed annotators who satisfy the following requirements to evaluate our results: (1) HITs approval rate greater than 95\%, (2) Location is one of Australia, Canada, New Zealand, United Kingdom, and the United States, (3) Lifetime number of HITs approved greater than 1000, following \citet{li2018towards}.
Figure~\ref{fig:5_mturk} shows the instructions and the interface for the human evaluation.
To mitigate the bias from the annotator, we randomly shuffle the order of the model and the corresponding response.
\subsection{Automatic Evaluation}
For automatic metrics, we calculate the metric for each case and take the average of those values.
When calculating BLEU, we use \texttt{sentence\_bleu} function in \texttt{nltk} python package~\cite{Loper02nltk:the}.

\section{Measuring Inference Time}
We measure how much time spend when the model generates the responses.
When generating the response, \textit{Blender 90M} takes $0.481$ seconds, and \textit{RetNRef}+\textit{CORGE} takes $0.523$ seconds per instance.
There is only an $8.7\%$ amount of inference time gap between \textit{Blender 90M} and \textit{RetNRef}+\textit{CORGE}.
This tells us that exemplar-based generation can significantly improve the quality of responses regarding appropriateness, informativeness, and diversity without increasing the amount of time to generate answers.
We test our model on NVIDIA DGX Station A100 with PyTorch 1.7.1, CUDA 11.0, CuDNN 8.0, and here we adopt the generation strategy we describe above.
When we measure the inference time, we only use a single GPU (NVIDIA A100 GPU, 40GB Memory), and the inference time is measured as the average inference time of 100 response generations.

\section{Additional Results}
We provide additional samples for the retrieved exemplar and the model response from the baselines and our models in Table~\ref{tab:6_additional_results}.

\end{document}